\documentclass[11pt,a4paper]{article}
\usepackage[hyperref]{acl2019}
\usepackage{times}
\usepackage{latexsym}
\usepackage{graphicx}
\usepackage{url}
\usepackage{algorithm}
\usepackage{algpseudocode}
\usepackage{amsmath}
\usepackage{amssymb}
\usepackage{multirow}

\aclfinalcopy 
\def\aclpaperid{467} 
\usepackage{xcolor}

\title{Coupling Retrieval and Meta-Learning for \\Context-Dependent Semantic Parsing}

\author{Daya Guo$^1$\thanks{\ \ \ Work done while this author was an intern at Microsoft Research.} , Duyu Tang$^2$, Nan Duan$^2$, Ming Zhou$^2$, and Jian Yin$^1$\\
$^1$ The School of Data and Computer Science, Sun Yat-sen University.\\
	Guangdong Key Laboratory of Big Data Analysis and Processing, Guangzhou, P.R.China\\
	$^2$ Microsoft Research Asia, Beijing, China\\
	{\tt \{guody5@mail2,issjyin@mail\}.sysu.edu.cn}\\
	{\tt \{dutang,nanduan,mingzhou\}@microsoft.com}\\	
}

\date{}

\begin{document}
\maketitle
\begin{abstract}
	In this paper, we present an approach to incorporate retrieved datapoints as supporting evidence for context-dependent semantic parsing, such as generating source code conditioned on the class environment. 
	Our approach naturally combines a retrieval model and a meta-learner, where the former learns to find similar datapoints from the training data, and the latter considers retrieved datapoints as a pseudo task for fast adaptation. 
	Specifically, our retriever is a context-aware encoder-decoder model with a latent variable which takes context environment into consideration, and 
	our meta-learner learns to utilize retrieved datapoints in a model-agnostic meta-learning paradigm for fast adaptation. 
	We conduct experiments on CONCODE and CSQA datasets, where the context refers to class environment in JAVA codes and conversational history, respectively.
	We use sequence-to-action model as the base semantic parser, which performs the state-of-the-art accuracy on both datasets.
	Results show that both the context-aware retriever and the meta-learning strategy improve accuracy, and our approach performs better than retrieve-and-edit baselines.
\end{abstract}

\section{Introduction}
Context-dependent semantic parsing aims to map a natural language utterance to a structural logical form (e.g. source code) conditioned on a given context (e.g. class environment) \cite{ling2016latent,long2016simpler,iyyer2017search,iyer2018mapping, suhr2018learning, Suhr:18situated}. 
Standard approaches typically learn a one-size-fits-all model on the entire training dataset, which is fed with each example individually in the training phase and makes predictions for each test example in the inference phase.
However, taking code generation as an example, programmers usually do not write codes from scratch in the real world. 
When they write a piece of code in a particular environment, they typically leverage past experience on writing or reading codes in the similar situation as a guidance.
Meanwhile, datapoints for a task may vary widely \cite{huang2018natural}, thus it is desirable to learn a ``personalized'' model for the target datapoint.
In this work, we study how to automatically retrieve similar datapoints in a context-dependent scenario and use them as the supporting evidence to facilitate semantic parsing.

\begin{figure}[h]
	\centering
	\includegraphics[width=.47\textwidth]{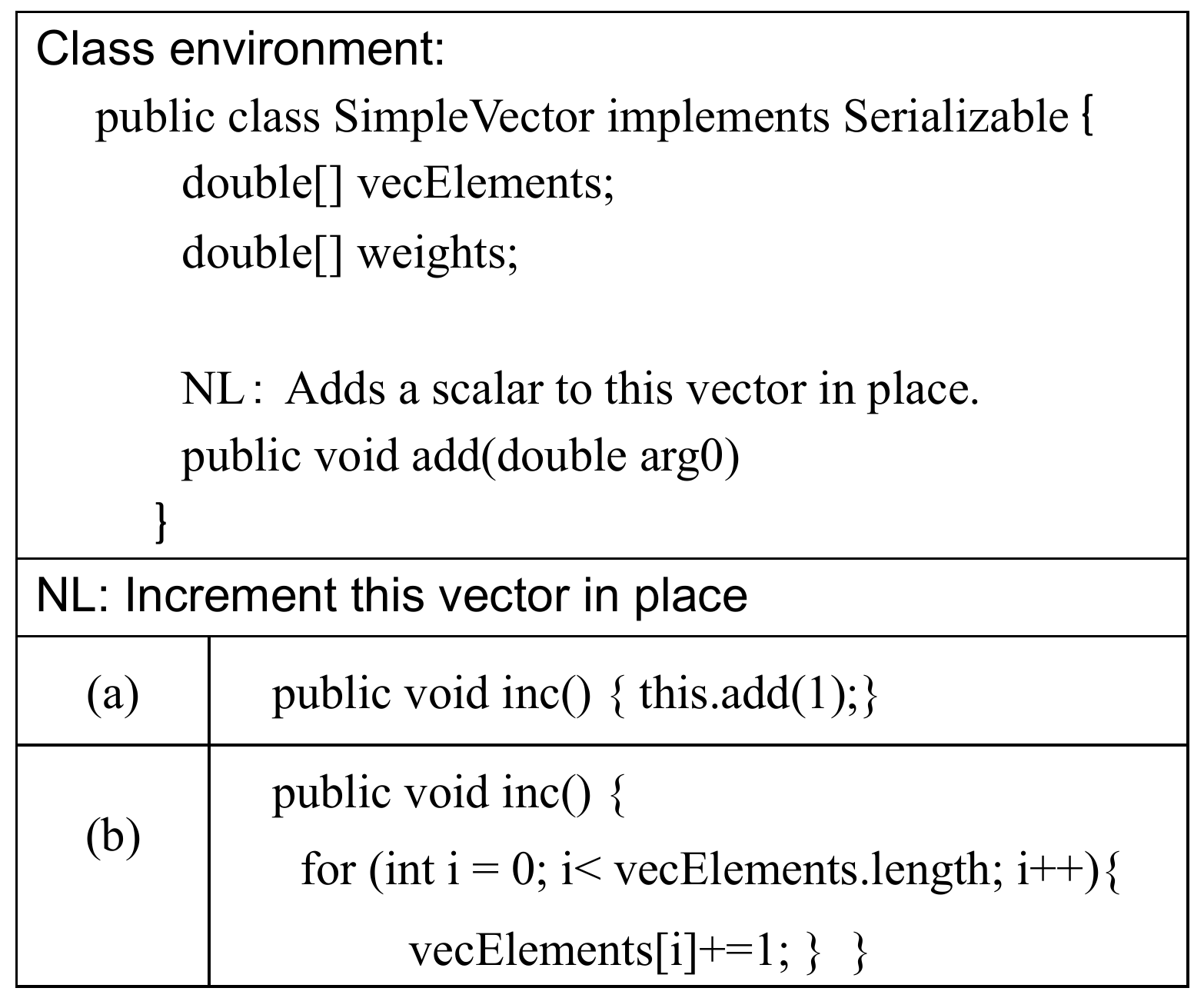}
	\caption{Code generation based on the class environment and a natural language documentation (NL). (a) shows a example of code generation by applying the class function $add()$, while (b) iterates the $vecElements$ array to increment each element.}
	\label{fig:example}
\end{figure}

There are recent attempts at exploiting retrieved examples to improve the generation of logical form and text. 
Retrieve-and-edit approaches \cite{hashimoto2018retrieve,huang2018dictionary,wu2018response,gu2017search} typically first use a context-independent retriever to find the most relevant datapoint, and then use it as an additional input of the editing model. 
However, a context-aware retriever is very important for the task 
of context-dependent semantic parsing. 
For examples, as shown in Figure \ref{fig:example}, class environment can help the retriever decide whether the desired code of \textit{``Increment this vector''} is generated by directly calling $add()$ or iterating the $vecElements$ array to increment each element. 
Furthermore, retrieve-and-edit approaches typically consider only one similar example to edit. 
In semantic parsing, the pattern of a structural output may come from different retrieved examples. 
There also exist works to utilize multiple examples to guide the semantic parser \cite{hayati2018retrieval,huang2018natural}, however, these approaches either use a heuristic way to exploit the retrieved logical form such as increasing the probability of actions \cite{hayati2018retrieval} 
or use a relevance function designed and learned based on expertise about the target logical form \cite{huang2018natural}.
When we consider the context environment, it's nontrivial to design a context-aware relevance function since the form of the context environment varies widely.

In this work, we propose to retrieve similar examples by taking into account the context environment, and use meta-learning to utilize retrieved examples to guide the generation of a logical form. 
Our retriever is a context-aware encoder-decoder model that takes context environment into consideration. Specially, the model is  based on the variational auto-encoder framework \cite{kingma2013auto,rezende2014stochastic}, which encodes a natural language utterance with the context environment into a latent variable that can produce the correct logical form. 
We adopt meta-learning framework \cite{finn2017model} to train a general semantic parser that can quickly adapt to a new (pseudo) task via  few-shot learning, where multiple retrieved examples are viewed as a support set of a pseudo task. Our approach naturally make use of multiple similar examples to guide the semantic parser in the current task. 





We evaluate our approach on CONCODE \cite{iyer2018mapping} and CSQA \cite{saha2018complex} datasets, where tasks are generating source code conditioned on the class environment in JAVA codes and answering conversational question over a knowledge graph conditioned on conversational history.
Results show that our approach achieves the state-of-the-art performances on both datasets.
We show that coupling retrieval and meta-learning performs better than two retrieve-and-edit baselines.
Further analysis show that both the context-aware retriever and the meta-learning strategy improve the performance.

\section{Task Definition and Datasets}
Context-dependent semantic parsing aims to map a natural language to a structural logical form conditioned on the context environment. 
In this section, we introduce two tasks we study, namely code generation and conversational question answering, and the datasets we use. 

\subsection{Context-dependent Code Generation}
Figure \ref{fig:example} shows a example of code generation. Given a natural language (NL) description  $x$, the goal aims to generate a source code $y$ conditioned on the class environment $c$. Formally, the class environment comprises two kinds of context: (1) class variables $v$ composed of variable names and their data type (e.g. {$double[\ ]\ vecElements$), and (2) class methods $m$, including method names with their return type (e.g. $void\ add()$). 
We conduct experiments on the CONCODE\footnote{\url{https://github.com/sriniiyer/concode}} dataset \cite{iyer2018mapping}. 
The dataset is built from about 33,000 public Java projects on Github that contains NL and codes together with class environment information.


\subsection{Conversational Question Answering}
This task aims to answer questions in conversations based on a knowledge base (KB). We tackle the problem in a context-dependent semantic parsing manner.
Specially, the task aims to map the question $x$ conditioned on conversational history $c$ into a logical form $y$, which will be executed on the KB to produce the answer. 
The conversational history refers to preceding questions $\{q_1, q_2, .., q_{i-1}\}$. In particular, we use the CSQA\footnote{\url{https://amritasaha1812.github.io/CSQA}} dataset \cite{saha2018complex} to develop our model and to conduct the experiments.  
The dataset is created based on Wikidata with 12.8M entities, including 152K/16K/28K dialogs for training/development/testing.

\section{Overview of the Approach}
We present our approach in this section, which first retrieves supporting datapoints from the training dataset using a context-aware retriever, and then considers retrieved datapoints as a pseudo task for fast adaptation in a model-agnostic meta-learning paradigm \cite{finn2017model}.
Figure \ref{fig:framwork} gives an overview of our approach.
\begin{figure}[t]
	\centering
	\includegraphics[width=.47\textwidth]{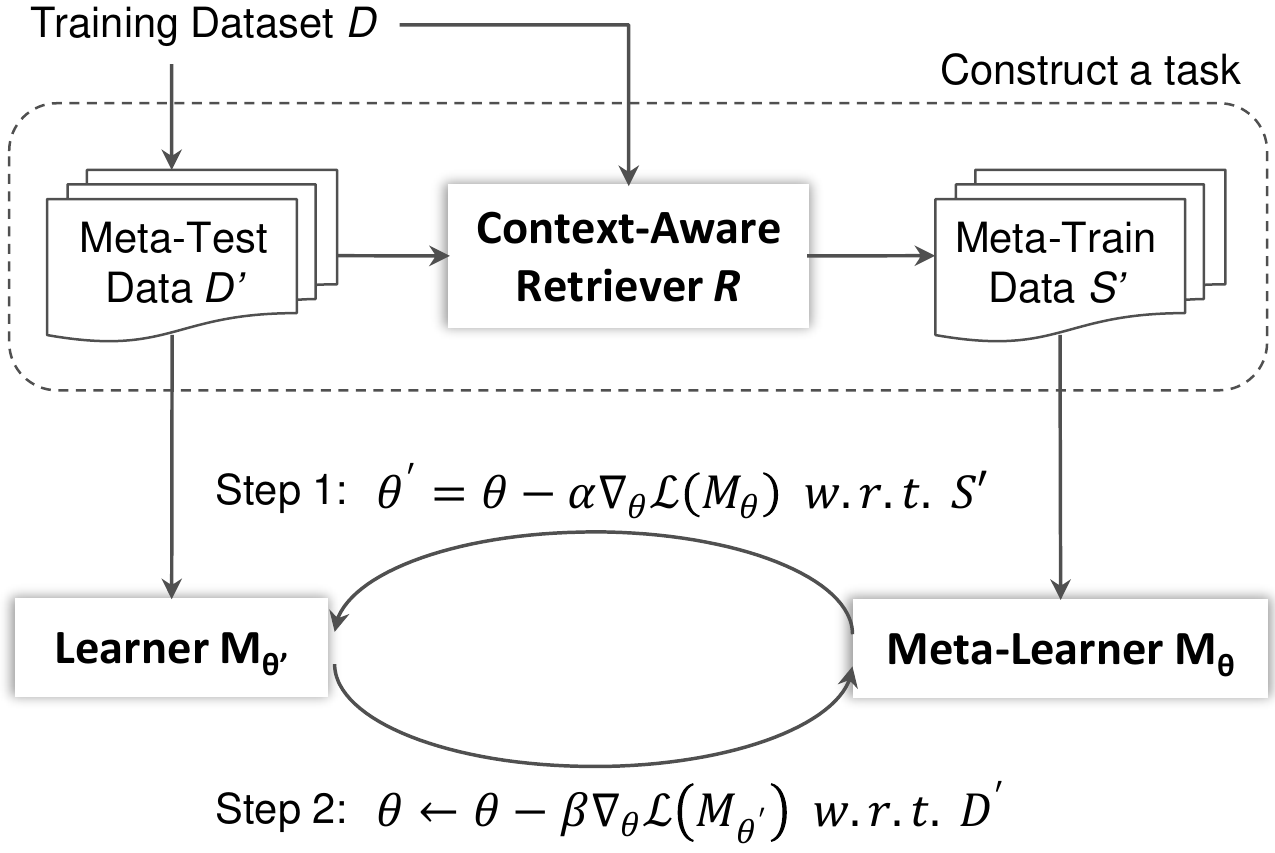}
	\caption{An overview of our approach that couples context-aware retriever and  meta-learning.}
	\label{fig:framwork}
\end{figure}
First, we sample a batch of examples $D'$ from the training dataset $D$. In meta-learning, there are two optimizing steps, namely the meta-train step (Step 1 in Figure \ref{fig:framwork}) that learns a task-specific learner $M_{\theta'}$ based on the current parameter $\theta$, and the meta-test step (Step 2 in Figure \ref{fig:framwork}) that updates the parameter $\theta$ based on the evaluation of $M_{\theta'}$.
In this work, $D'$ is used for meta-test process, and retrieved examples $S'$ from the context-aware retriever are used for meta-training.
In the inference phase, We consider the prediction of each test example as a new task, given retrieved examples from the training data as the supporting evidence. 
Instead of applying the general model $M_{\theta'}$ directly, retrieved examples are used to update the model, and the updated model will be used to make predictions.
The approach is summarized in Algorithm \ref{alg:Framwork}.

The details about the context-aware retriever and the semantic parser model will be introduced in Sections \ref{sec:retriever} and Section \ref{sec:learner}, respectively.
\begin{algorithm}[htb]
	\caption{Retrieval-MAML}
	\label{alg:Framwork}
	\begin{algorithmic}[1]
		\Require \
		Training dataset $D=(x^{(j)},c^{(j)},y^{(j)})$, step size  $\alpha$ and $\beta$
		\Ensure \
		Meta-learner $M$
		\State Training a context-aware retriever $R$ using $D$. 
		\State For each example $d$, we obtain a support set $S^d$ retrieved by $R$
		\State Randomly initialize $\theta$ for $M$
		
		\While{not done}
		\State {Sample a batch of examples $D'$ from $D$ as test examples, and $S'=\bigcup _{d\in D'} S^{d}$ are viewed as training examples }
		\State {Evaluate $\nabla_\theta\mathcal{L}(M_\theta)$ using $S'$, and compute adapted parameters with gradient descent: $\theta'=\theta-\alpha\nabla_\theta\mathcal{L}(M_\theta)$}
		\State  {Update $\theta\leftarrow\theta-\beta\nabla_\theta\mathcal{L}(M_{\theta'})$ using $D'$ for meta-update}
		\EndWhile
		
	\end{algorithmic}
\end{algorithm}

\section{Context-Aware Retriever}
\label{sec:retriever}
In this section, we present the model architecture of our context-aware retriever, the way to use the model to retrieve similar examples using a distance metric in the latent space, and how to effectively train the model.


%


\subsection{Model Architecture}
Figure \ref{fig:retriever} illustrates an overview of the retrieval model in the task of generating source code. 
Following \citet{hashimoto2018retrieve}, our retriever is a encoder-decoder model based on the variational autoencoder framework, which encodes a natural language $x$ with the context environment $c$ into a latent variable $z$ that can predict the output $y$.

\begin{figure*}[t]
	\centering
	\includegraphics[width=1\textwidth]{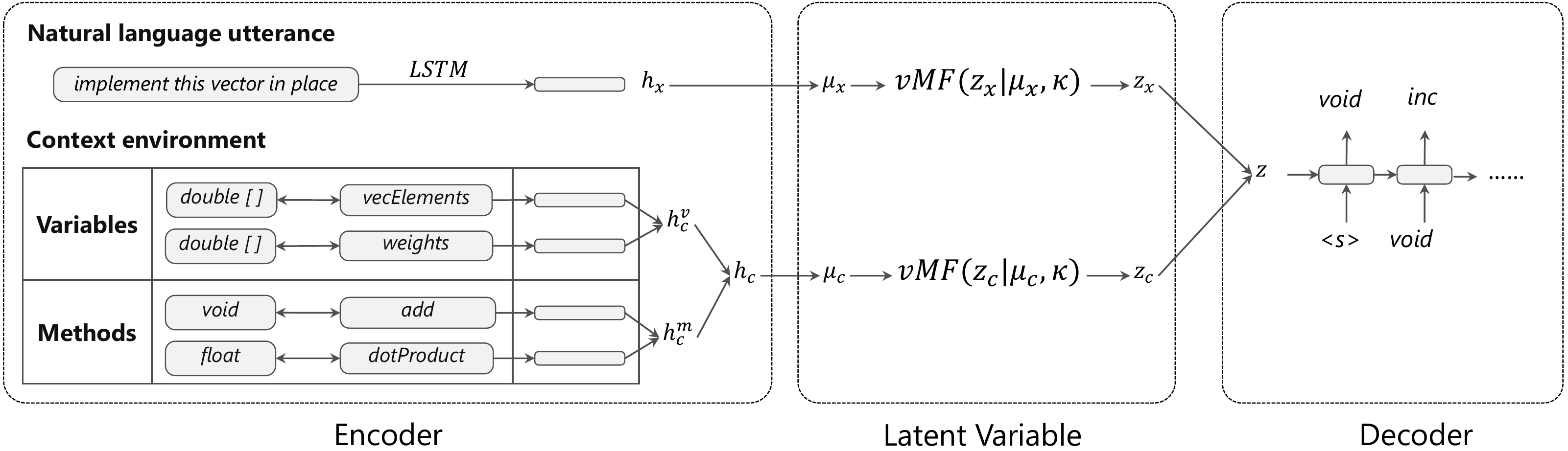}
	\caption{An overview of our context-dependent retriever.}
	\label{fig:retriever}
\end{figure*}

\paragraph{Encoder}
We use bidirectional RNNs with LSTM \cite{hochreiter1997long} as encoders to compute the representation $h_x$ of the natural language $x$ and the representation $h_c$ of the context environment $c$, where $h_x$ is the hidden states of the NL encoder at the last token and details about $h_c$ for CONCODE and CSQA datasets are provided in the appendix A.

\paragraph{Latent variable}
We have two latent variables, one ($z_x$) is for the current utterance and another ($z_c$) is for the context. 
We use the  concatenation of $z_x$ and $z_c$ as the embedding of the natural language with the context, namely $z=[z_x;z_c]$. 

We describe how to map the the natural language $x$ into a latent variable $z_x$ here. The calculation of $z_c$ is analogous to $z_x$. 
Following \cite{hashimoto2018retrieve}, we choose $z_x$ to be a von Mises-Fisher (vMF) distribution over unit vectors centered on $\mu_x$, where both $z_x$ and $\mu_x$ are unit vectors, and $Z_{\kappa}$ is a normalization constant depending only on constant $\kappa$ and the dimension $d$ of $z_x$. The $\mu_x$ is calculated by a linear layer followed by a activation function, and the input is $h_x$. 
\begin{equation}\label{equa:z_x}
p(z_x|x)= vMF(z_x;\mu_x,\kappa)=Z^{-1}_{\kappa}e^{({\kappa}{\mu_x}^Tz_x)}
\end{equation}

Other distributions such as the Gaussian distribution can also be used to represent latent variables, but we choose the vMF in this paper since the KV-divergence is proportional to the squared Euclidean distance between their respective direction vectors $\mu$ with the same $\kappa$ and $d$. The property will be used in the next section.

\paragraph{Decoder}
At the decoding, we first sample $z$ from $p(z|x,c)$ using the re-parametrization trick \cite{kingma2014auto-encoding}, and then use an additional linear layer over $z$ to obtain the initial hidden state of the decoder. 
We use LSTM as the decoder. At each time-step $t$, the current hidden state $s_t$ of the decoder is used to predict a word from the vocabulary. 
In order to ensure that the target $y$ is only inferred by the latent variable $z$, we don't incorporate attention or copying mechanism. The strategy is also used in \citet{hashimoto2018retrieve}.

\subsection{Retrieve Examples}
We use KL-divergence as the distance metric to retrieve similar examples in the latent space. 
In particular, the KL divergence between two vMF distributions with the same concentration parameter $\kappa$ is calculated as follows, where $\mu_1,\mu_2 \in \mathbb{R}^{d-1}$ and $C_\kappa$ is calculated as Equation \ref{C_kappal}.

\begin{equation}\label{equa:d-kl}
KL(vMF(\mu_1,\kappa)||vMF(\mu_2,\kappa))=C_\kappa||\mu_1-\mu_2||^2_2
\end{equation} 
$I_d$ stands for the modified Bessel function of the first kind at order $d$. Since $C_\kappa$ only depends on $\kappa$ and $d$, the KL divergence is proportional to the squared Euclidean distance between their respective direction vectors $\mu$ with the same $\kappa$ and $d$. More details about the proof of this proposition can be found in \cite{hashimoto2018retrieve}.

\begin{equation}\label{C_kappal}
C_\kappa=\kappa\frac{I_{d/2}(\kappa)}{2I_{d/2-1}(\kappa)}
\end{equation} 
Given two examples $(x,c)$ and $(x',c')$, the KL divergence between their distributions of latent variables (i.e. $p(z|x,c)$ and $p(z|x',c')$) is equivalent to the distance calculated as given in Equation \ref{equa:distance}. The retriever will find top-$K$ nearest examples according to the distance. 
\vspace{-0.1cm}
\begin{equation}\label{equa:distance}
\begin{aligned}
distance&=KL(p(z|x,c)||p(z|x',c'))\\
&=KL(p(z_x|x)||p(z_x|x')) \\
&\quad +KL(p(z_c|c)||p(z_c|c'))\\
&=C_\kappa(||\mu_x-\mu'_x||^2_2+||\mu_c-\mu'_c||^2_2)
\end{aligned}
\end{equation}

\subsection{Training}
Our entire approach corresponds to the following generative process. Given a example $(x,c)$, we first use the retriever $p_{ret}(S|x,c)$ to find similar examples $S$ as a support set and then generate an output $y$ by a meta-learner model $p_{m}(y|x,c,S)$ based on $S$. Therefore, the probability distribution over targets $y$ is formulated in Equation \ref{p_y}.
\begin{equation}\label{p_y}
p(y|x,c)=\sum_{S\subset D}p_{m}(y|x,c,S)p_{ret}(S|x,c)
\end{equation}
A basic idea for learning the retriever might be maximizing the marginal likelihood by jointly learning, but it is computationally intractable. Instead, we train the retriever in isolation, assuming that semantic parser provides the true conditional distribution over the target $y$ given context $c$ and retrieved examples $S$ under the joint distribution $p_{ret}(S|x,c)p_{data}(x,c,y)$. 
Then, we optimize a lower bound for the marginal likelihood under this semantic parser \cite{hashimoto2018retrieve}, which decomposes the reconstruction term and the KL divergence as follows. 
\begin{align}\label{equa:KL}
\notag
logp(y|x,c)\geq E_{z\sim p(z|x,c)}logp(y|z)  \\
-E_{\{({x',c')\}} \sim p_{ret}}KL(p(z|x,c)||p(z|x',c'))
\end{align}  
According to Equation \ref{equa:distance}, the upper bound of $KL(p(z|x,c)||p(z|x',c'))$ is $8C_\kappa$. Therefore, we can maximize the this worst-case lower bound, where $C_\kappa$ is constant in our case. This lower bound objective is analogous to the recently proposed hyperspherical variational autoencoder \cite{davidson2018hyperspherical,xu2018spherical}.
\begin{equation}\label{equa:KL1}
logp(y|x,c)\geq E_{z\sim p(z|x,c)}logp(y|z)-8C_\kappa
\end{equation} 
Thus, we optimize the context-aware retriever by maximizing $E_{z\sim p(z|x,c)}logp(y|z)$.


\section{Semantic Parser}
\label{sec:learner}
Recently, sequence-to-action models \cite{yin2017syntactic,P18-1071,iyer2018mapping,guo2018dialog} have achieved strong performance in semantic parsing, which consider
the generation of a logical form as the prediction of a sequence of actions (e.g. derivation rules in a defined grammar). 
We use two context-dependent sequence-to-action models \cite{iyer2018mapping,guo2018dialog} as the base semantic parsers, both of which take a natural language with the context environment as the input and outputs an action sequence. 
Both models achieve state-of-the-art on CONCODE and CSQA datasets.

In the task of code generation,  
the JAVA abstract grammar contains a set of production rules composed of an non-terminal and multiple symbols (e.g. $Statement \rightarrow return \ Expression$). 
We represent a source code as an Abstract Syntax Tree (AST) by applying several production rules \cite{aho2007compilers}. 
The sequence of production rules applied to generate an AST is viewed as an action sequence $a_1,...,a_n$, where an action $a$ refers to a production rule. 
To access to the context environment, we introduce several special actions. For examples, two actions $IdentifierNT \rightarrow ClassMethod$ and $ClassMethod \rightarrow constant$ are used to invoke class methods. The former action means that the identifier comes from class methods, and the latter is an action used for  instantiating $ClassMethod$ by a copying mechanism \cite{gu2016incorporating}. 
In the task of conversational question answering, we convert the logical form into a action sequence using the similar grammar defined in \cite{guo2018dialog}. 

%
Specially, a bidirectional LSTM takes a source sentence as the input, and feeds the concatenation of both ends as the initial state of the decoder. The decoder has another LSTM to generate an action sequence in a sequential way.
At each time-step $t$, the decoder calculates the current hidden state $s^{dec}_{t}$ as Equation \ref{equa:update4}, where $s^{dec}_{t-1}$ is the last hidden state, $n_t$ is the current non-terminal to be expanded, and $y_{t-1}$ is the previously predicted action. $p_{n_t}$ and $s_{n_t}$ are the parent action and the hidden state of the decoder respectively, which produce the current non-terminal. If the previously predicted action is an instantiated action, the embedding $y_{t-1}$ is the representation of the selected constant.
\begin{equation}\label{equa:update4}
s^{dec}_t= LSTM(s^{dec}_{t-1},[n_t;y_{t-1};p_{n_t};s_{n_t}])
\end{equation}

In order to generate a valid logical form, the model incorporates an action-constrained grammar to filter illegitimate actions. 
An action is legitimate if its left-hand non-terminal is the same as the current non-terminal to be expanded.
Let us denote the set of legitimate actions at the time step $t$ as $A_t=\{a_1,...,a_N\}$. The probability distribution over the set is calculated as Equation 
\ref{equa:action distribution}, 
where $v_i$ is the one-hot indicator vector for $a_i$, $W_a$ is model parameter, and $a_{<t}$ stands for the preceding actions of the $t$-th time step.
\begin{equation}\label{equa:action distribution}
p(a_i|a_{<t},x)=\frac{exp(v_i^TW_as^{dec}_t)}{\sum_{a_j\in A_t}exp(v_j^TW_as^{dec}_t)}
\end{equation} 
For instantiated actions (e.g. $ClassMethod \rightarrow constant$), the probability of a constant $m$ being instantiated at time-step $t$ is calculated as Equation \ref{equa:Instantiation function}, where $W$ is model parameter, $v_m$ is the embedding of the constant.
\begin{equation}\label{equa:Instantiation function}
p(m|a_{<t},x)=\frac{exp(v_m^Ttanh(Ws^{dec}_t))}{\sum_{m'}exp(v_{{m}'}^Ttanh(Ws^{dec}_t))}
\end{equation}

Please see more details about the model hyperparameters in the Appendix B.

  \begin{table*}[t]
 	\centering
 	\begin{tabular}{l|cc|cc}
 		\hline
 		\multirow{2}{*}{Methods}& \multicolumn{2}{c|}{Dev} & \multicolumn{2}{c}{Test} \\
 		& {Exact} & {BLEU} & {Exact} & {BLEU}\\
 		\hline
 		\hline
 		\multicolumn{5}{c}{Retrieval ONLY} \\
 		\hline
 		\hline
 		TFIDF &1.25&17.78&1.50&19.73\\
 		Context-independent Retrieval &0.85&19.63&0.80&21.98\\
 		Context-dependent Retrieval & 1.30& 21.21& 1.00& 24.94\\
 		\hline
 		\hline
 		\multicolumn{5}{c}{Parsing-based methods without retrieved examples} \\
 		\hline
 		\hline
 		Seq2Seq   &2.90&21.00&3.20&23.51\\
 		Seq2Prod \cite{yin2017syntactic} &5.55&21.00&6.65&21.29\\
 		 \citet{iyer2018mapping}  &7.05&21.28&8.60&22.11\\
 		\bf Seq2Action    &\bf 7.75&\bf 22.47&\bf 9.15&\bf 23.34\\
 		\hline
 		\hline
 		\multicolumn{5}{c}{Parsing-based methods with retrieved examples} \\
 		\hline
 		\hline
 		Seq2Action+Edit vector (Context-independent Retrieval)  &6.6&21.27&7.90&22.51\\
 		Seq2Action+Edit vector (Context-aware Retrieval) &7.75&20.69&9.20&22.68\\
 		Seq2Action+Retrieve-and-edit (Context-independent Retrieval)  &5.55&21.27&7.05&22.74\\
 		Seq2Action+Retrieve-and-edit (Context-aware Retrieval)  &7.55&22.20&9.30&23.95\\
 		Seq2Action+MAML (Context-independent Retrieval) &9.15 &21.48&9.85&23.22\\
 		Seq2Action+MAML (Context-aware Retrieval, w/o finetune) &8.30& 21.27&10.30&24.12 \\		
 		\bf Seq2Action+MAML (Context-aware Retrieval) & \bf 8.45& \bf 21.32 & \bf 10.50& \bf 24.40\\	
 		\hline
 	\end{tabular}
 	\caption{Performance of different approaches on the CONCODE dataset.}
 	\label{table:concode}

 \end{table*}
 
 \section{Experiment}

 \subsection{Model Comparisons on CONCODE}
Table \ref{table:concode} reports results of different approaches on the CONCODE dataset.
We use Exact match accuracy as the major evaluation metric, which measure whether the generated program is exactly correct. Following \citet{iyer2018mapping}, we also report BLEU-4 score \cite{papineni2002bleu} between the reference and generated code as a reference.
These approaches are divided into three groups. The first group is retrieval ONLY, which directly returns the top-ranked retrieved example. 
The second group report numbers of existing systems and our base model Seq2Action, all of which do not use retrieved examples. 
Models in the last group utilize retrieved examples.

From the first group, we can see that directly using the retrieved output has extremely low Exact score since of mismatching environment variables and methods, which means that its meaning is incorrect. Yet, the BLEU score is acceptable, which means that some constituents might be useful.
In the second group, we compare parsing based methods without retrieved examples. As we can see, our \textbf{Seq2Action} model outperforms others models, resulting in the state-of-the-art accuracy without using retrieved examples.
In the third group, we implement two retrieval-augmented methods for comparison. 
\textbf{Retrieve-and-edit} uses a copying mechanism to copy tokens from the retrieved example \cite{hashimoto2018retrieve}.
\textbf{Edit vector} calculates an edit vector by considering lexical differences between a prototype context and current context, and uses the edit vector as an extra feature  \cite{wu2018response}. 
We can see that applying the MAML framework to the Seq2Action model achieves a gain of 1.35\% exact match accuracy.
Results also show that our context-aware retriever performs better than the context-independent retriever in various settings.
 
 \begin{table*}[t]
 	\centering
 		
 		\centering
 		\begin{tabular}{l|c|c|c|c|c|c}
 			\hline
 			\multirow{2}{*}{Methods} &HRED & \multirow{2}{*}{D2A}&\multirow{2}{*}{S2A}&S2A &S2A  &S2A\\
 			 & +KVmem& & & +EditVec& +RAndE& +MAML\\
 			\hline
 			Question Type&\multicolumn{6}{c}{F1}\\
 			\hline
 			Simple Question (Direct)&13.64\%&91.41\%&92.01\%&91.95\%&92.08\%&\bf 92.66\% \\
 			Simple Question (Co-referenced)&7.26\%&69.83\%&71.40\%& 72.94\%&\bf73.19\%&71.18\%\\
 			Simple Question (Ellipsis)&9.95\%&81.98\%&81.75\%& 83.31\%&\bf 84.61\%&82.21\%\\
 			Logical Reasoning (All)&8.33\%&43.62\%&42.00\%&43.85\% &41.83\%&\bf 44.34\%\\
 			Quantitative Reasoning (All)&0.96\%& 50.25\%&45.37\%&46.93\%&42.64\%&\bf 50.30\% \\
 			Comparative Reasoning (All)&2.96\%&44.20\%&41.51\%&43.96\% &44.46\%&\bf 48.13\%\\
 			Clarification&16.35\%&18.31\%&18.9\%&18.42\%& 18.70\%&\bf19.12\% \\
 			\hline
 			Question Type&\multicolumn{6}{c}{Accuracy} \\
 			\hline
 			Verification (Boolean) &21.04\%&45.05\%& 51.17\%&47.81\%&\bf 55.00\%&50.16\% \\
 			Quantitative Reasoning (Count)  &12.13\%&40.94\%&46.01\%&44.67\%&43.07\%&\bf 46.43\% \\	
Comparative Reasoning (Count)  &5.67\%&17.78\%&16.52\%&17.52\%&16.43\%&\bf 18.91\% \\	
 			\hline
 		\end{tabular}
 	\caption{Performance of different approaches on the CSQA dataset.}
 	\label{table:csqa}
 \end{table*}
 
 \subsection{Model Comparisons on CSQA}
We follow the experiment protocol of \citet{guo2018dialog}.
To make the comparison clearer,
we use F1 score as evaluation metrics for questions whose answers are sets of entities.  Accuracy is used to measure the performance for questions which produce boolean and numerical answers. 
Table \ref{table:csqa} shows the results of different methods on the CSQA dataset. More detailed numbers are provided in the appendix C.
\textbf{HRED+KVmem} \cite{saha2018complex} is a encoder-decoder model with key-value memory network \cite{D16-1147} to directly produce answers. \textbf{D2A} \cite{guo2018dialog} is a sequence-to-action model described in Section \ref{sec:learner}. Since the dataset does not provide annotated action sequence for each question, we follow \cite{guo2018dialog} to use a breadth-first-search algorithm to obtain action sequences that lead to correct answers. 
However, some of action sequences are spurious \cite{P17-1097}, in the sense they do not represent the meaning of questions but get the correct answers. We use retrieved examples by our context-aware model to filter out spurious action sequences. We choose the most similar action sequence to retrieved action sequences, measured by editing distance. We denote the model learned in this way as \textbf{S2A}. 

Table \ref{table:csqa} shows that filtering out spurious action sequences brings about 5\% point improvement on boolean and Quantitative Reasoning (Count) questions. 
Results also show that applying the \mbox{MAML} framework performs better than both retrieve-and-edit approaches, namely RAndE \cite{hashimoto2018retrieve} and EditVec \cite{wu2018response},
on the majority of question types, especially on complex questions.

\begin{figure}[t]
	\centering
	\includegraphics[width=.48\textwidth]{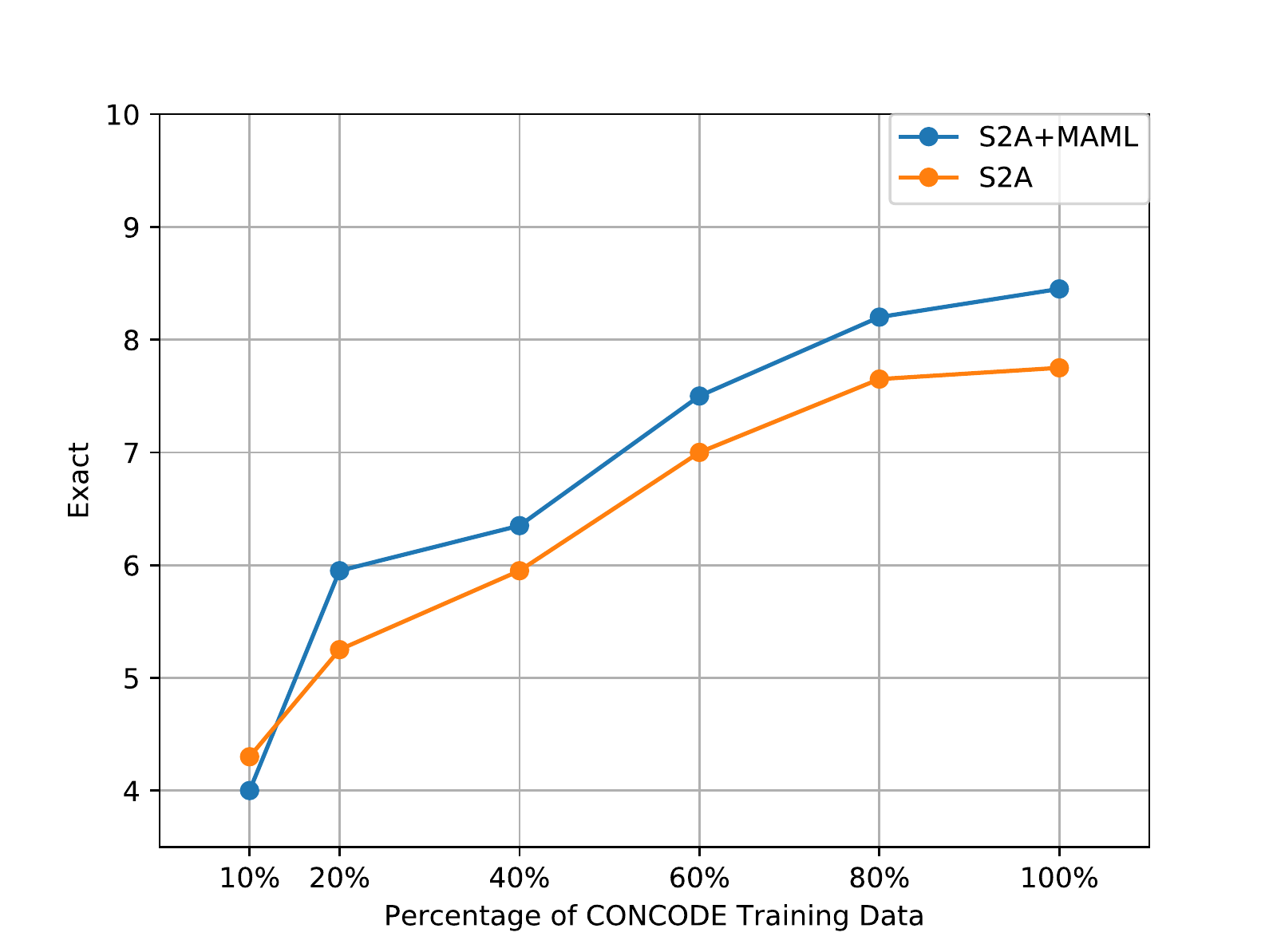}
	\caption{Comparison between S2A and S2A+MAML with different portions of supervised data.}
	\label{fig:datasize}
\end{figure} 

\begin{figure}[t]
	\centering
	\includegraphics[width=.47\textwidth]{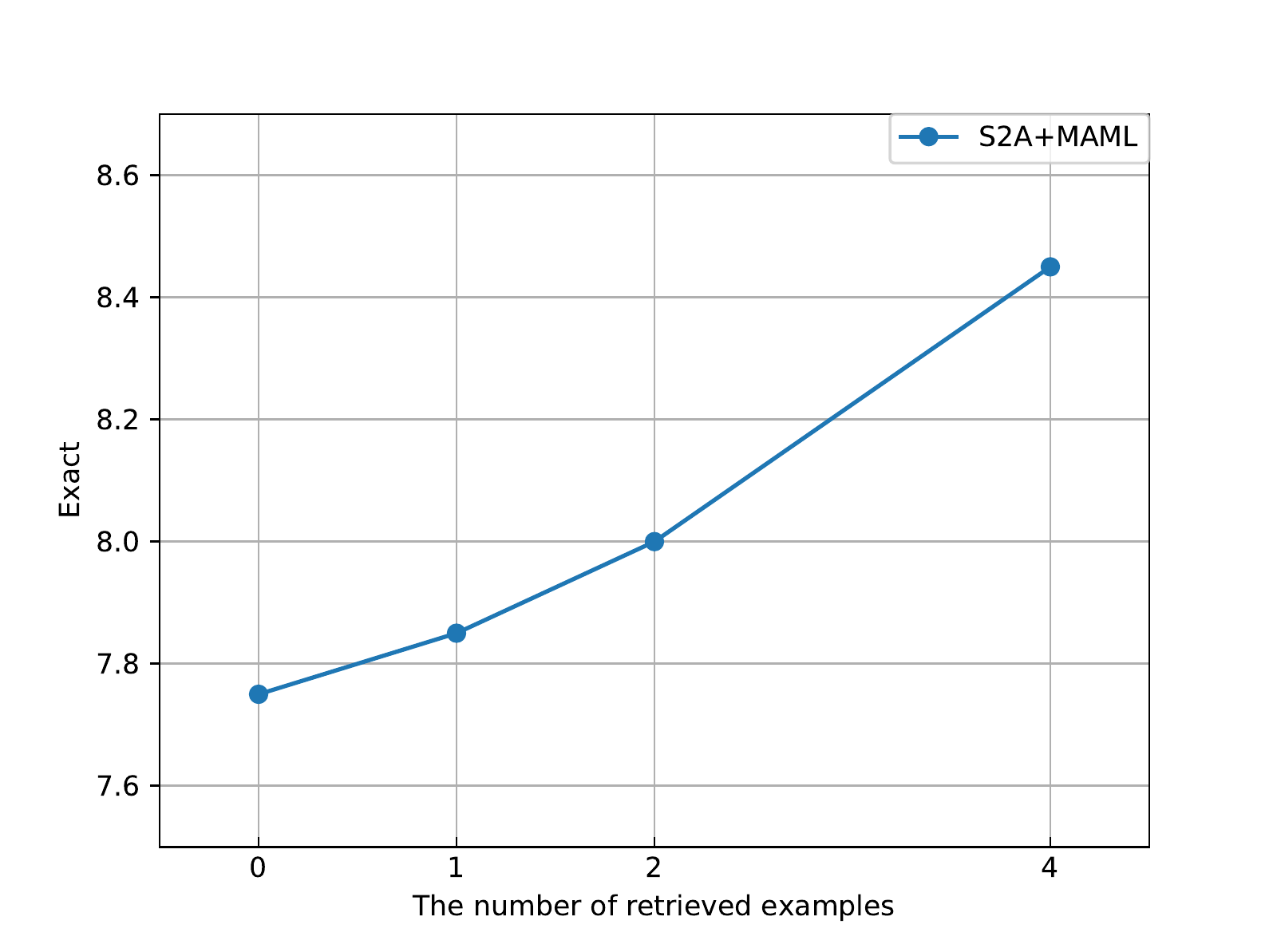}
	\caption{S2A+MAML with different number of retrieved examples on the CONCODE devset.}
	\label{fig:k}
\end{figure}

\subsection{Model Analysis}
We study how the amount of training dataset and retrieved examples impacts the overall performance on the CONCODE. 
From Figure \ref{fig:datasize}, we can see that S2A+MAML performs better than S2A in when $>$20\% supervised datapoints are available to retrieve from.
From Figure \ref{fig:k}, we can see that the accuracy increases as the number of retrieved examples expands. This is consistent with our intuition that the performance of the semantic parser is improved by utilizing multiple retrieved examples, since the pattern of a logical form may come from different retrieved examples. 
We did not try larger number of retrieved examples due to the memory limit of our GPU device.
However, excessive retrieved examples may introduce noise, which \mbox{hurts} the performance of the semantic parser. 
Therefore, we need to choose the appropriate amount of retrieved examples.

\begin{figure*}[t]
	\centering
	\includegraphics[width=1\textwidth]{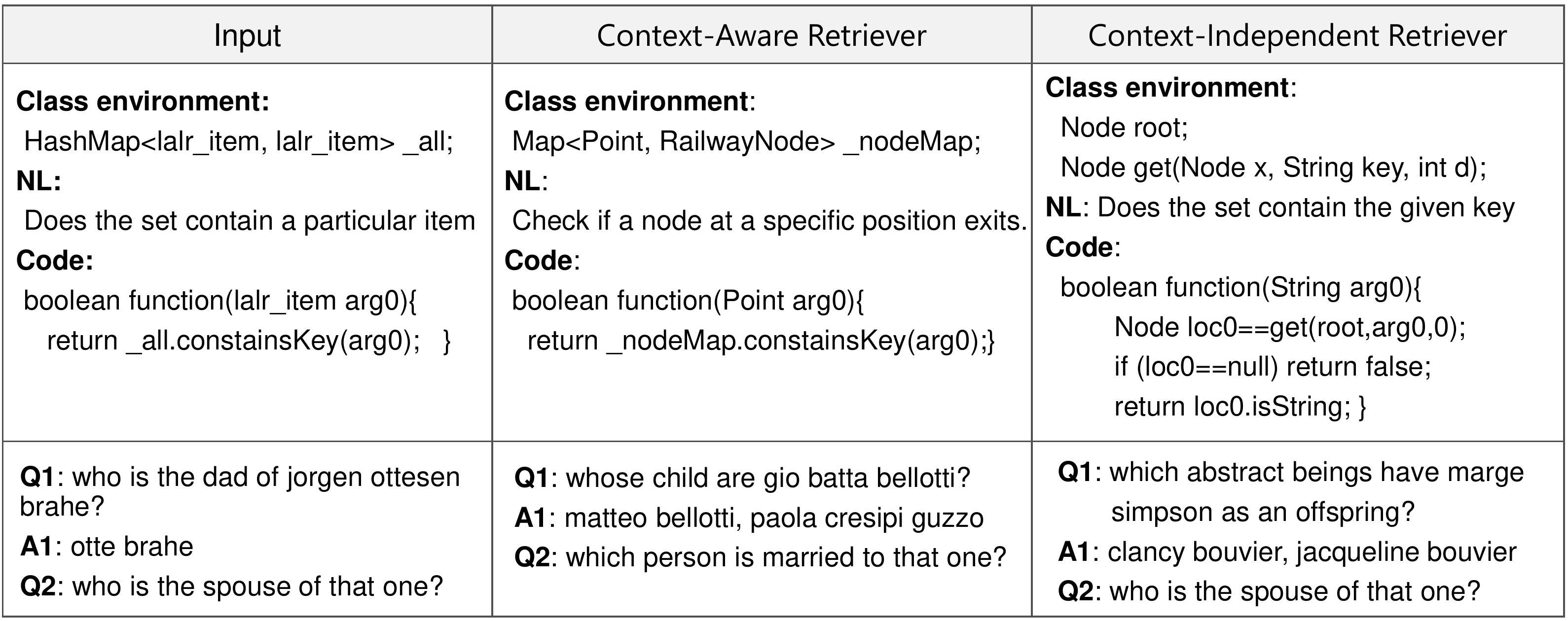}
	\caption{Examples from the CONCODE dataset (first row) and the CSQA dataset (second row). The retrieved examples found by context-aware retriever (center panels) and context-independent retriever (right panels) follow the input (left panels).}
	\label{fig:cases}

\end{figure*}

 \subsection{Case Study}
 We give a case study to illustrate the retrieved results by our context-aware retriever, with a comparison to the context-independent retriever. Results are given in Figure \ref{fig:cases}. We can see that our retriever can capture semantic content to retrieve. For examples, in the second row, the current question (i.g. \textit{``who is the spouse of that one''}) has the same semantic as that one of the context-aware retrieved example (i.g. \textit{``which person is married to that one''}), which demonstrates that our retriever learns the semantic of \textit{``spouse''} and \textit{``married''} in the retrieval process. 
Comparing with the context-independent retriever, incorporating the \mbox{context} environment can improve the performance of the retrieval. 
Taking the first row as a example, although the NL of the input (i.g. \textit{``Does the set contain a particular item''}) have similar semantic to that one of the context-independent retriever (i.g. \textit{``Does the set contain the given key''}), source codes differ greatly because the types of the sets are different (i.g. $HashMap$ and $Node$ respectively). 
Our context-aware retriever can find the example with similar source code by considering  their context environment (both $HashMap$ and $Map$ have same $constainsKey$ function).

\subsection{Error Analysis}
We analyze a randomly selected set of wrongly predicted 100 instances on the CONCODE dataset. 
We observe that 44\% examples do not correctly copy class members, among which the majority of them lack information about class member (e.g. the effect of the class method $get$). This problem might be mitigated by encoding source codes of class methods or incorporating descriptions of class members. 
24\% examples fail to invoke functions of the library and member class (e.g. a model is required to know there exit a $size()$ function in List class to invoke $list.size()$). 
A potential direction to mitigate the problem is incorporate definitions of the external or system classes, which requires an updated version of the dataset.
Among the other 32\% examples, the major problem is that some of retrieved examples are incorrect.
 Incorporating more signal to measure the usefulness of retrieved examples might alleviate this problem.

\section{Related work}
Semantic parsing is a fundamental problem in NLP that maps natural languages to logical forms of their underlying meaning, including variable-free logic \cite{zelle1995comparative,clarke2010driving}, lambda calculus \cite{Zettlemoyer05,kushman2013using}, dependency-based compositional semantics \cite{liang2011learning,berant2013semantic}, and database queries \cite{iyer-EtAl:2017:Long,zhong2017seq2sql,P17-1003}. 
Recently, context-dependent semantic parsing has drawn plenty of attention \cite{long2016simpler,iyyer2017search,iyer2018mapping, suhr2018learning, Suhr:18situated}, where the generation of logical forms is conditioned on the context environment. 
 
Neural encoder-decoder models have proved effective in semantic parsing \cite{neelakantan2015neural,dong-lapata:2016:P16-1,yin2017syntactic,D18-1190}.  One direction is to employ sequence-to-sequence model by modeling semantic parsing as a sentence to logical form  translation problem \cite{dong-lapata:2016:P16-1,jia2016data,ling2016latent,xiao2016sequence}.
However, regarding logical form as a sequence could not guarantee the grammatical correctness of the generated output. Sequence-to-Action approaches  \cite{yin2017syntactic,krishnamurthy-dasigi-gardner:2017:EMNLP2017,iyyer2017search,P18-1071} treat semantic parsing as the prediction of a action sequence that can construct logical forms, which not only guarantee the grammatical correctness of outputs, but also leverage the strength of sequence-to-sequence model in learning sequential transformations.

Recently, there are recent attempts at exploiting retrieved examples to improve the generation of logical forms.
\citet{hashimoto2018retrieve} propose a retrieve-and-edit approach, including an encoder-decoder based retrieval model learnt in a task-dependent way without relying on a hand-crafted metric, and an editing model with a copying mechanism to replicate tokens from the retrieved example.  
\citet{hayati2018retrieval} increase the probability of actions that can derive the retrieved subtrees.
\citet{huang2018natural} also use MAML and treat each example as a new task. 
The relevance function for retrieving examples is based on the predicated type of the SQL query and the question length.
Different from these three works, we focus on context-dependent semantic parsing, and our context-aware retriever is learned from the dataset without the help of a hand-craft relevant function.
Different from \cite{hashimoto2018retrieve}, 
our approach naturally make use of multiple similar examples to improve the semantic parser. 

Retrieval-augmented models have also been studied in text generation \cite{gu2017search,huang2018dictionary,guu2018generating,wu2018response}. \citet{gu2017search} use the retrieved sentence pairs as extra inputs to the NMT model. \citet{wu2018response} calculate an edit vector by considering lexical difference between a prototype context and current context, which is used as extra features. 

  

\section{Conclusion}
In this paper, we present an approach which combines a context-aware retrieval model and model-agnostic meta-learning (MAML)  to utilize multiple retrieved examples for context-dependent semantic parsing.
 We show that both context-aware retriever and MAML are useful on CONCODE and CSQA datasets.
 Our approach achieves the state-of-the-art performances and outperforms two retrieve-and-edit baselines. 

\section*{Acknowledgments}

This work is supported by the National Key R\&D Program of China (2018YFB1004404), Key R\&D Program of Guangdong Province (2018B010107005), National Natural Science Foundation of China (U1711262, U1401256, U1501252, U1611264, U1711261). Jian Yin is the corresponding author.

\bibliography{acl2019}
\bibliographystyle{acl_natbib}

\appendix

\section{Encoders for Context-Aware Retriever}
To obtain the representation of the context environment $h_c$, we first use a 2-step Bi-LSTM to encode the variable to get the contextual representation of class variables $h^{i}_v$. The representations of class methods $h^{i}_m$ are also obtained in the same way. Usually, identifiers are composed of multiple words, such as $vecElements$. We split them based on camel-casing and encode them to get corresponding embedding by a Bi-LSTM.
Finally, the representation of the context environment $h_c$ is calculated 
by average pooling over the vectors.

Different from code generation, the context of conversational question answering refers to previous questions $\{q_1, q_2, .., q_{i-1}\}$. Therefore, we use a bidirectional LSTM to encode them to get the representation $\{h_{q_1}, h_{q_2}, .., h_{q_{i-1}}\}$. The representation of the context environment $h_c$ is calculated by average pooling over the vectors.

\begin{table*}[t]	
	\centering	
	\begin{small}
		
		\centering
		\begin{tabular}{l|cc|cc|cc}
			\hline
			Methods & \multicolumn{2}{c|}{HRED+KVmem}& \multicolumn{2}{c|}{D2A}&\multicolumn{2}{c}{S2A}\\
			\hline
			Question Type&Recall&Precision&Recall&Precision&Recall&Precision \\
			\hline
			Overall&18.40\%&6.30\%&64.04\%&61.76\%&64.86\%&62.51\%\\
			Simple Question (Direct)&33.30\%&8.58\%&93.67\%&89.26\%&93.19\%&90.86\% \\
			Simple Question (Co-referenced)&12.67\%&5.09\%&71.31\%&68.41\%&72.96\%&69.91\%\\
			Simple Question (Ellipsis)&17.30\%&6.98\%&86.58\%&77.85\% &85.24\%&78.54\%\\
			Logical Reasoning (All)&15.11\%&5.75\%&42.49\%&44.82\% &44.30\%&39.93\%\\
			Quantitative Reasoning (All)&0.91\%&1.01\%&48.59\%&52.03\%&45.98\%&44.77\% \\
			Comparative Reasoning (All)&2.11\%&4.97\%&44.73\%&43.69\%&43.23\%&39.93\% \\
			Clarification&25.09\%&12.13\%&19.36\%&17.36\%&19.84\%&18.04\% \\
			\hline
			Question Type&\multicolumn{2}{c|}{Accuracy}&\multicolumn{2}{c|}{Accuracy}&\multicolumn{2}{c}{Accuracy} \\
			\hline
			Verification (Boolean) &\multicolumn{2}{c|}{21.04\%}&\multicolumn{2}{c|}{45.05\%}&\multicolumn{2}{c}{51.17\%} \\
			Quantitative Reasoning (Count)  &\multicolumn{2}{c|}{12.13\%}&\multicolumn{2}{c|}{40.94\%}&\multicolumn{2}{c}{46.01\%} \\	
			Comparative Reasoning (Count)  &\multicolumn{2}{c|}{8.67\%}&\multicolumn{2}{c|}{17.78\%}&\multicolumn{2}{c}{16.52\%} \\	
			\hline
		\end{tabular}
		
	\end{small}
	\caption{Results of methods without utilizing retrieved examples on the CSQA dataset.}
	\label{table:basemethod}
\end{table*}

\begin{table*}[t]
	\centering

	\begin{small}
		\centering
		\begin{tabular}{l|cc|cc|cc}
			\hline
			Methods & \multicolumn{2}{c|}{S2A+EditVec}& \multicolumn{2}{c|}{S2A+RAndE}&\multicolumn{2}{c}{S2A+MAML}\\
			\hline
			Question Type&Recall&Precision&Recall&Precision&Recall&Precision \\
			\hline
			Overall&65.51\%&63.45\%&65.54\%&63.12\%&65.23\%&63.02\%\\
			Simple Question (Direct)&93.47\%&90.48\%&93.72\%&90.50\%&94.43\%&90.95\% \\
			Simple Question (Co-referenced)&74.11\%&71.81\%&74.47\%&71.96\%&72.72\%&69.70\% \\
			Simple Question (Ellipsis)&87.01\%&79.91\%&88.06\%&81.42\%&85.89\%&78.84\% \\
			Logical Reasoning (All)&42.21\%&45.63\%&40.55\%&43.20\%&42.59\%&46.23\% \\
			Quantitative Reasoning (All)&48.26\%&45.67\%&45.44\%&40.17\%&50.77\%&49.83\% \\
			Comparative Reasoning (All)&46.60\%&41.60\%&47.08\%&42.11\%&48.32\%&47.95\% \\
			Clarification&19.30\%&17.61\%&19.81\%&17.71\%&20.01\%&18.31\% \\
			\hline
			Question Type&\multicolumn{2}{c|}{Accuracy}&\multicolumn{2}{c|}{Accuracy}&\multicolumn{2}{c}{Accuracy} \\
			\hline
			Verification (Boolean) &\multicolumn{2}{c|}{47.81\%}&\multicolumn{2}{c|}{55.00\%}&\multicolumn{2}{c}{50.16\%} \\
			Quantitative Reasoning (Count)  &\multicolumn{2}{c|}{44.67\%}&\multicolumn{2}{c|}{43.07\%}&\multicolumn{2}{c}{46.43\%} \\	
			Comparative Reasoning (Count)  &\multicolumn{2}{c|}{17.52\%}&\multicolumn{2}{c|}{16.43\%}&\multicolumn{2}{c}{18.91\%} \\	
			\hline
		\end{tabular}
	\end{small}
	\caption{Results of methods with utilizing retrieved examples on the CSQA dataset.}
	\label{table:retrieved_method}
\end{table*}

\section{Model Training}
For the context-aware retriever on both experiment, we set the dimension of the word embedding as 300. The encoder is a 2-layer bidirectional LSTM with hidden states of size 300, and the decoder is a 4-layer unidirectional LSTM with hidden states of size 300. We use dropout with a rate of 0.5, which is applied to the inputs of RNN. We set the dimension of latent variable and $\kappa$ as 600 and 500, respectively. Model parameters are initialized with uniform distribution, and updated with the Adam method. We set the learning rate as 0.0001 and the batch size as 20. We tune hyper parameters and perform early stopping on the development set.

The hyperparameters of encoder and decoder are the same as the context-aware retriever. We follow \cite{huang2018natural} to train the meta-learner without back-propagating to second order gradients. The number of retrieved examples for the CONCODE dataset and CSQA dataset are 4 and 1 respectively, which are tuned on the development sets. We set the step size $\alpha$ of task update as 0.001 for both experiment. On the CONCODE dataset, the learning rate $\beta$ is 0.0002 and the test-batch size is 10, while the CSQA dataset are 0.001 and 32 respectively.

\section{Results on CSQA}
Here, we provide more detailed numbers about the performance of different approaches on the CSQA dataset. Precision and recall are used as evaluation metrics for questions whose answers are sets of entities. Accuracy is used to measure the performance for questions which produce boolean and numerical answers. Table \ref{table:basemethod} shows the results of methods without utilizing retrieved examples, and Table \ref{table:retrieved_method} shows the results of retrieval-augmented methods.

\end{document}